\documentclass[journal]{vgtc}                     

\usepackage{misc/cool}

\onlineid{1572}

\vgtccategory{Research}

\vgtcpapertype{algorithm/technique}

\title{Visualizing Topological Importance: A Class-Driven Approach}

\author{%
  Yu Qin, Brittany Terese Fasy, Carola Wenk, and Brian Summa}
  
\authorfooter{
\item
 Yu Qin is with Tulane University. E-mail: yqin2@tulane.edu.
\item
 Brittany Terese Fasy is with Montana State University. E-mail: brittany.fasy@montana.edu.
\item
 Carola Wenk is with Tulane University. E-mail: cwenk@tulane.edu.
\item
 Brian Summa is with Tulane University. E-mail: bsumma@tulane.edu.
}

\abstract{

This paper presents the first approach to visualize the importance of topological features that define classes of data.  Topological features, with their ability to abstract the fundamental structure of complex data, are an integral component of visualization and analysis pipelines.  Although not all topological features present in data are of equal importance. To date, the default definition of feature importance is often assumed and fixed. This work shows how proven explainable deep learning approaches can be adapted for use in topological classification. In doing so, it provides the first technique that illuminates what topological structures are important in each dataset in regards to their class label.  In particular, the approach uses a learned metric classifier with a density estimator of the points of a persistence diagram as input. This metric learns how to reweigh this density such that classification accuracy is high. By extracting this weight, an importance field on persistent point density can be created.  This provides an intuitive representation of persistence point importance that can be used to drive new visualizations. This work provides two examples:  Visualization on each diagram directly and, in the case of sublevel set filtrations on images, directly on the images themselves. This work highlights real-world examples of this approach visualizing the important topological features in graph, 3D shape, and medical image data. }


\teaser{
  \centering
  \includegraphics{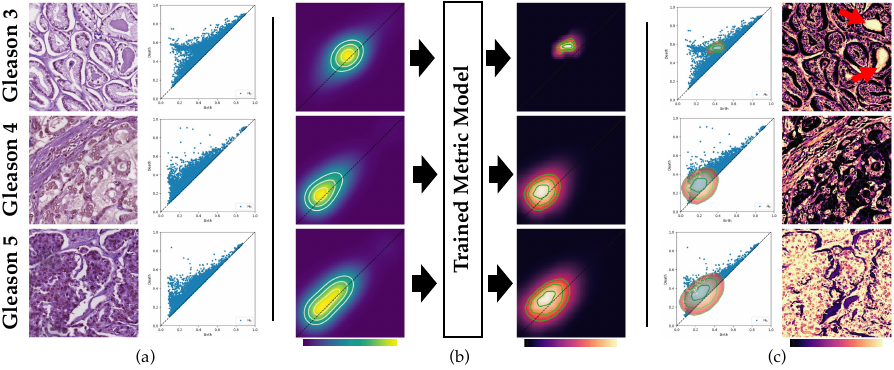}
  \caption{(a) A common task in topological data analysis: extracting a persistence diagram of topological features. In this case, features are based on the sublevel set filtration of pathology images with class labels (Gleason grade) that define the progression of prostate cancer~\cite{lawson2019persistent}. Knowing which features are important for each class is commonly an educated guess with the lifetime of a feature (persistence) often assumed to define importance. (b) Our approach, based on a learned metric classifier, takes as input the unweighted density of persistence points and reweighs this density based on what best defines a class. This allows us to build a field of importance for regions of a diagram. (c) This importance field can be used to create visualizations to illuminate which features drive a classification. For example, it can highlight what points are important directly in a diagram or, in the case of sublevel set filtrations, visualize the important structure directly in an image. Consider that a hallmark of prostate cancer is gland degeneration as the disease progresses.  Calcifications (red arrow) are only present in well-structured glands and are highlighted as important structures for Gleason 3, an earlier stage of the disease.}
  \label{fig:teaser}
}

\graphicspath{{figs/}{figures/}{pictures/}{images/}{./}{misc/}} 

\usepackage{tabu}                      
\usepackage{booktabs}                  
\usepackage{lipsum}                    
\usepackage{mwe}                       

\usepackage{mathptmx}                  

\begin{document}


\firstsection{Introduction}

\maketitle

\label{sec:introduction}

Topological data analysis (TDA)~\cite{edelsbrunner2010computational} is a crucial component of many data analysis and visual analytics pipelines. Features from TDA, extracted using persistent
homology~\cite{edelsbrunner2000topological}, contour
trees~\cite{carr2003computing}, Reeb
graphs~\cite{biasotti2008reeb,pascucci2007robust}, and Morse(--Smale)
complexes~\cite{de2015morse,gyulassy2008practical}, provide important abstractions of data structure in applications ranging from physics~\cite{bremer2010interactive,gyulassy2014stability,maljovec2016topology,kasten2011two} and chemistry~\cite{bhatia2018topoms,gunther2014characterizing} to medicine~\cite{lee2014hole,lee2011discriminative,shnier2019persistent,lawson2019persistent}, to name a few. As a natural consequence of the importance of these features, researchers often want to use data topology to drive analysis tasks such as classification.

Although, there is little intuition about which topological features are important to define a class, or to what degree. For example, it is commonly assumed that the persistence (lifetime in function value) of a feature is a good weight for importance, as low-persistence features' ephemeral lifetimes are often attributed to noise. But, contrary to this assumption, work~\cite{patrangenaru2019challenges,bubenik2020persistent} has shown that low-persistence features are more important for some types of data. Therefore, the current practice of determining which features to target and which to discount is to assume persistence, make an educated guess, or, worse, determine the correct weights for features as the result of a trial-and-error process. In response to the diversity of datasets, it is necessary to develop a visualization approaches that can aid users in understanding which features define important structures in a dataset.

In this work, we introduce an approach to provide such visualizations. The core of our work is to use proven, explainable deep learning methods from computer vision on unweighted, vectorized density estimators of the points in persistence diagrams.  Our metric learning approach automatically learns the regional importance of topological features in a diagram and the weights on densities that are necessary for accurate classification. Rather than assume importance weights (persistence) or find them through trial-and-error, we learn them. As a result of using explainable deep learning, our approach provides an importance field over a diagram that allows TDA researchers for the first time to determine which features define a class and which do not matter. As an initial step towards interpreting topological features, understanding the importance field across a diagram has dual benefits. It enhances both the field of TDA-based machine learning and TDA-based visualization by encoding more meaningful topological information.
Using this field, new visualizations can be designed to illuminate the critical features but also challenge any preconceived assumptions about fundamental structure in data. 
 For example, as our results will show, the commonly assumed single, uniform weighting strategy on diagram points is insufficient as importance varies by both class and dataset. While using deep learning for classification with interpretation is not new, this work is the first to apply such an approach to topological features and use said result to visualize topological importance.

This work has the following novel contributions:

\begin{itemize}
\item A field over the space of a dataset's persistence diagram that highlights regions of importance in defining a class;
\item An approach that utilizes this field to visualize feature importance directly for zero-dimensional features of a sublevel-set filtration of a scalar field;
\item A deep metric learning approach for classification using topological features that outperforms the accuracy of the state-of-the-art topological-based methods; and
\item Examples of our visualization approach highlighting, for the first time, the importance of topological
    features for classes of graph, shape, and medical data.
\end{itemize}

\begin{figure}
   \centering
    \includegraphics[width=\columnwidth]{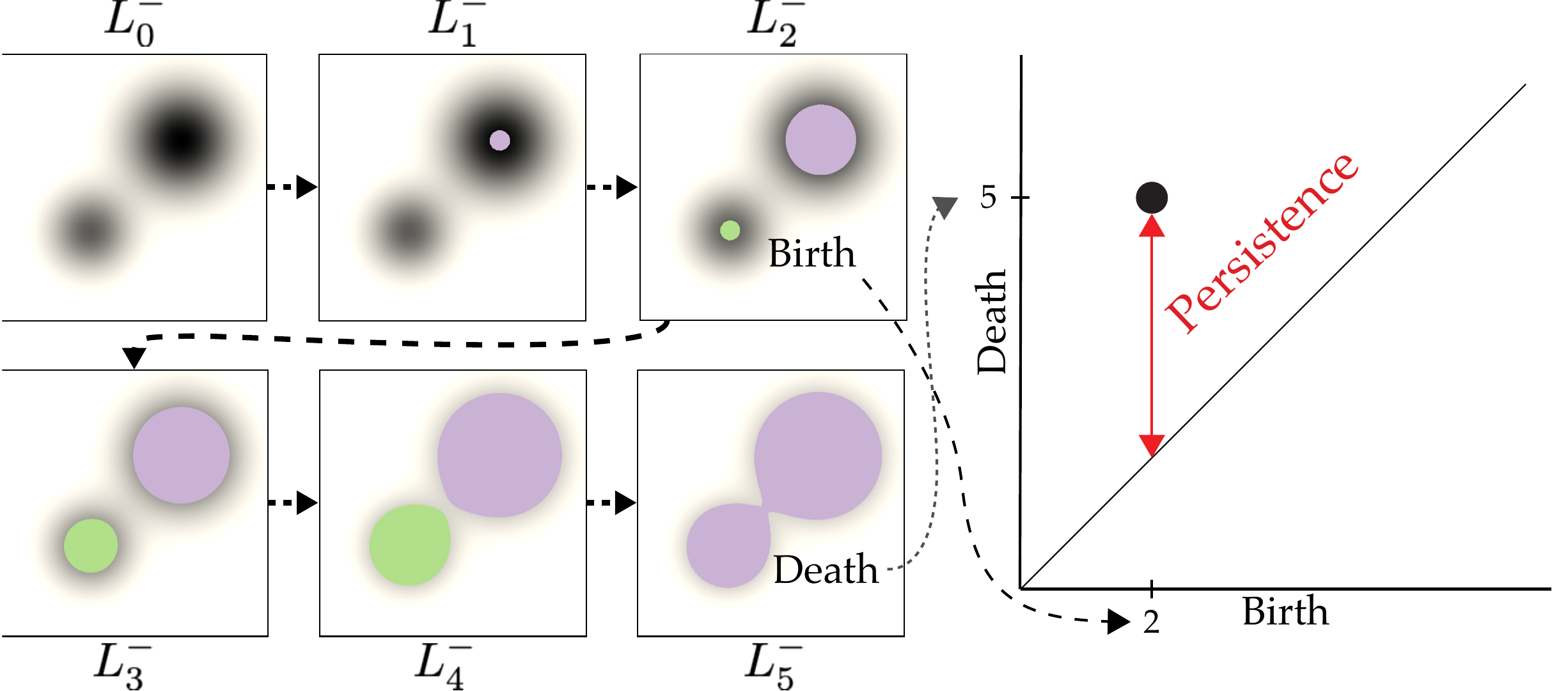}
        \caption{The progression of a sublevel-set ($L_{i}^{-}$) of a scalar
        field for increasing function values ($i$).  The green feature
        is born at the minimum introduced at $2$, and dies when it merges with an older
        feature (shown in purple) at $5$. The birth and death are represented as a point in the
        0D persistence diagram $(2,5)$.  The lifetime ($5-2=3$) of this
        feature is its persistence.
        }\label{fig:persistence}
        \vspace{-14pt}
\end{figure}

\section{Background and Related Work}
\label{sec:bg}

In this section, we begin with a brief introduction to an abstraction of topological features widely used in TDA, persistence diagrams. We then detail how these diagrams are extended to encode a richer set of features. For more details on this concept, we refer the reader to~\cite{edelsbrunner2010computational}.
Next, we discuss the weight functions on topological features and the need to learn these weights, vectorizations of persistence diagrams, and a brief overview of the approaches to visualize topological features.

\subsection{Topological Features and Persistence Diagrams}
\label{sec:pd}

Homology is a concept from algebraic topology that describes the
{\em holes} (connected components, cycles, voids, etc.) of a topological space.
Notationally, for each integer~$k$, we let $H_k(X)$ denote the $k$-th homology
group of a domain, $X$; see~\cite{hatcher,munkres2018elements} for details. For our purposes, we
use~$\Z_2$ coefficients, and so, the $k$-dimensional homology groups are vector
spaces that describe the $k$-dimensional holes of $X$.

A {\em filtration} is an ordered family of topological
spaces, connected by inclusion maps between them. For example,
if we let $X_i:=\{x\in X | f(x) < \alpha\}$ w.r.t. $\alpha \in \R$ denote sublevel set of $f$, we get a
nested sequence of topological spaces,~$X_1 \subseteq X_2 \subseteq \cdots \subseteq
X_n = X$. The inclusion~$X_i \hookrightarrow X_j$ for $i < j$ induces a linear map between the homology
groups~$H_k(X_i) \to H_k(X_j)$ on the corresponding $k$-th homology.
The two most common
filtrations are the sublevel sets of scalar functions (e.g., for image data)
or the evolution of a Vietoris--Rips
complex for unstructured data (e.g., for point clouds)~\cite{vietoris1927hoheren,hausmann1995vietoris}.

\begin{figure}[tb]
  \centering
    \includegraphics[width=1\linewidth]{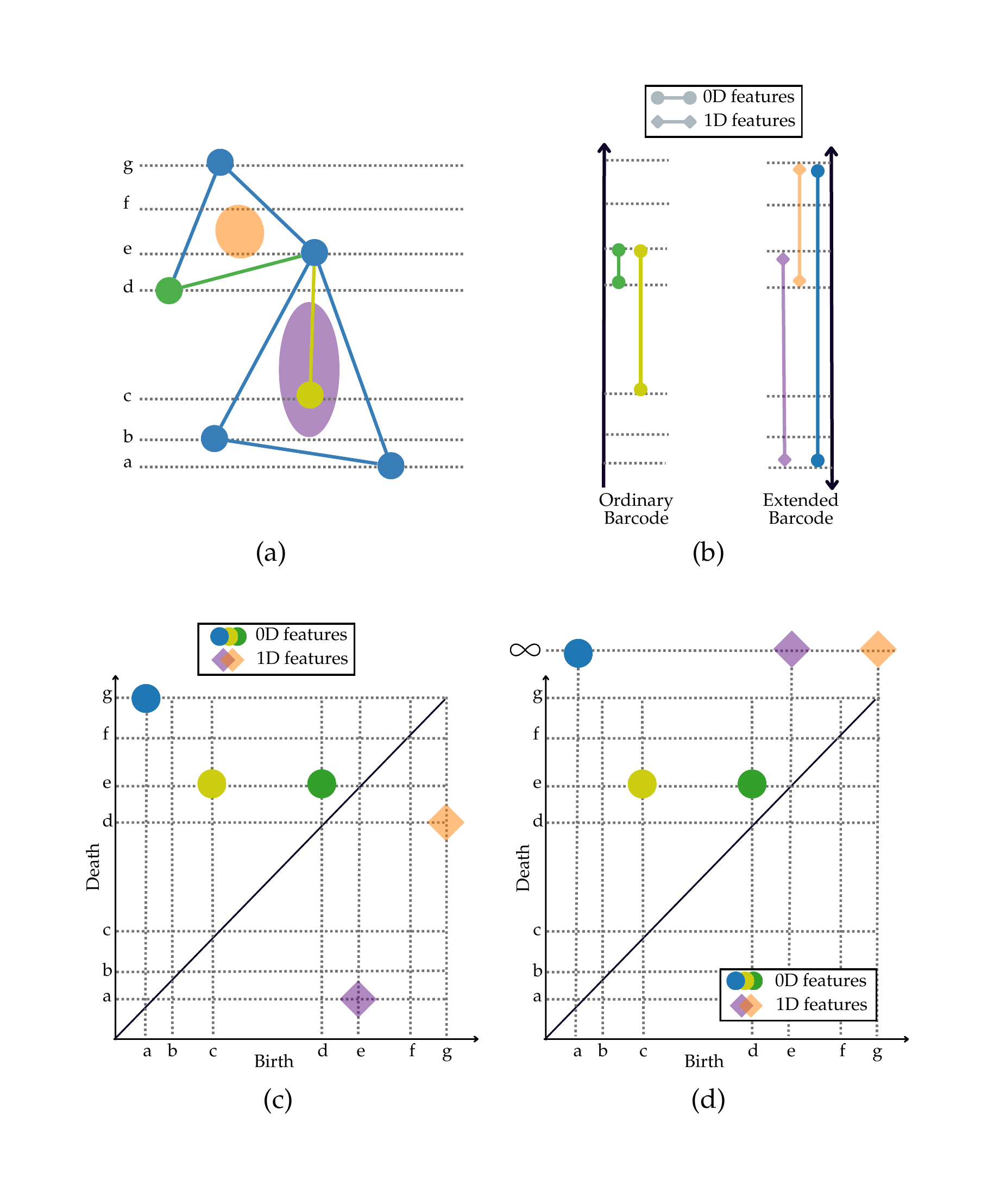}
        \vspace{-24pt}
    \caption{An illustration of an extended persistence diagram. (a) A graph with height filtration, where each node is associated with a filter value. (b) The ordinary and extended barcode.\protect\footnotemark ~ (c) The persistence diagram with extended filtration. (d) The persistence diagram with standard filtration. The extended persistence diagram highlights the effectiveness of the extended filtration function in capturing the additional topological information beyond the standard filtration.}
    \label{fig:pd}
            \vspace{-14pt}
\end{figure}
\footnotetext{The barcode in the extended filtration comprises three categories: ordinary, relative, and extended. For simplicity, we omit the relative barcode in this example since it is empty.}

Persistent homology~\cite{edelsbrunner2010computational}, roughly,
encodes the lifetime of a homological feature in this nested sequence. Homology provides a static description of topology, while persistent homology captures topology evolution over multiple scales through filtration and tracks changes in homology groups.
This is accomplished
by recording where a feature first appears ({\em birth}) and where it is
subsumed by an older feature ({\em death}).
Plotting this lifetime in~$\R^2$ (birth as the~$x$-coordinate and death as
the $y$-coordinate) gives a
{\em persistence diagram}. The diagram $D$ is composed of a set of points in the plane, where each point $(b,d)$ represents a feature. The feature corresponds to a $k$-dimensional homological structure that is created at the filtration value $X_b$ and destroyed at the filtration value $X_d$. In the case of a sublevel set filtration of a scalar field, these coordinates
are always function values of critical points. For example,
births for zero-dimensional (0D) features ($H_0$, i.e., connected components) occur
at local minima.
We call the difference between the birth time ($b$) and the death time ($d$), $|d-b|$, the
\emph{persistence} of the feature.  See~\figref{persistence}.

\paragraph{Extended persistence.} In the sublevel set filtration, the homology group of every topological space is
captured by going upward in function value. However, this filtration may be
insufficient in some contexts to encode the topology of domain $X$. For example,
let $X$ be a graph, in the case where $X$ contains cycles, the homology group of $X$ remains unchanged as the cycles never die. 

To address this limitation, an \textit{extended persistence diagram}~\cite{cohen2009extending} is proposed using an extended filtration. This approach ensures that every feature that appears in the space eventually disappears. We use relative homology theory and consider both upward and downward directions. Specifically, we compute the homology group going upward and the relative homology group coming back down. This results in paired births and deaths, where every feature that appears in the space eventually dies, and all births are paired with corresponding deaths.



In persistent homology, the extended filtration distinguishes between three categories of topological features: ordinary features that are born and die going upward, relative features that are born and die going downward, and extended features that are born going upward and die coming downward. 

This approach is best illustrated with a simple example. See \figref{pd} (a) with a graph with a scalar height function on the nodes.
First, we compute the persistence diagram using standard filtration by going upward. The corresponding topological features with finite lifetime under this filtration are defined as ordinary features, which capture two 0D features (i.e., connected components). Specifically, one feature is born at height $c$ and dies at $e$, while the other is born at height $d$ and dies at $e$. These two features are represented by the yellow and green lines in the ordinary barcode of \figref{pd} (b).

Additionally, three topological features are born and never die under this filtration, namely one 0D feature born at height $a$ and two 1D features born at height $e$ and $g$. We utilize relative homology theory to pair the death time of these topological features. Intuitively, if such feature is also created by going downward, then the corresponding time denotes the death time. This is because the downward-created feature represents a relative death time with respect to the upward-created feature that disappears. In the extended barcode of \figref{pd} (b), there is a 0D feature that is born at height $a$ going upward and dies at height $f$ coming downward (shown in blue line). Additionally, there are two unpaired 1D features: one is born at height $e$ going upward and dies at height $b$ going downward, while the other is born at height $g$ going upward and dies at height $d$ going downward (shown in purple and orange). Note that 1D features are born by going "up" and die by going "down." Therefore their birth time is larger than their death time. These features are encoded below the diagonal of the persistence diagram, as represented by the purple and orange diamonds in \figref{pd} (c).

Comparing with the persistence diagram under standard filtration in \figref{pd} (d), we observe that the extended persistence diagram in \figref{pd} (c) captures additional topological features. Specifically, the extended persistence diagram pairs three 0D and 1D topological features not paired in the standard filtration.
\paragraph{Wasserstein Distance.} The classic distance between persistence diagrams is the $p$-Wasserstein
distance~\cite{cohen2005stability}. At a high level, given two diagrams, this distance accumulates the cost of optimal point-wise matching between points of two diagrams. The diagonals of
the persistence diagrams are also viewed as having an infinite number of points. As low persistence points are close to the diagonal, they do not significantly add to the accumulation when not matched.  High persistence features are far from the diagonal and therefore incur a steeper penalty when they do not have a good match. Therefore, this distance naturally encodes persistence as a measure of importance.


\subsection{Weighting Topological Features}

Weighting topological features is essential in extracting more meaningful and relevant information from complex topological structures. Traditionally, the weight function is defined as the persistence of a feature, but as mentioned, persistence may not always be the most appropriate weight. Moreover, uniformly weighting all data does not account for any variance in the importance of topological features with respect to a dataset or class label. For instance,  Hofer et al.~\cite{hofer2017deep}  also
noticed that the weight function of a persistence diagram should not be pre-fixed
(i.e., weighting based on persistence). Similarly, both Harish et al.~\cite{doraiswamy2013topological} and Hamish et al.~\cite{carr2004simplifying} proposed methods that enable users to interactively define the importance of topological features. However, these methods require prior domain knowledge and do not integrate with any learning approaches.
Zhao et
al.~\cite{zhao2019learning} also has shown a real-world scenario in the atomic
configurations of molecules where low persistence features are most important
and, therefore, should be given a larger weight. Finally, Riihimäki and Licón-Saláiz~\cite{riihimaki2019metrics}, also highlighted the significance of low persistence features in topological persistence in their design of a contour metrics for topological features.


\subsection{Persistence Images}

In order to utilize topological features for downstream tasks, such as machine learning, it is necessary to transform them into vector representations. To accomplish this, several methods have been proposed that convert topological features into vectors~\cite{adams2017persistence,bubenik2020persistent,kusano2016persistence,carriere2017sliced,berry2020functional}.
One such vectorization is a {\em persistence image}, which is used by our approach.

Given a persistence diagram $D$ in birth-death $(b,d)$ coordinates. Let $T:\R^2 \rightarrow \R$ be the linear transformation: $T(b,d) = (b,d-b)$, and let $T(D)$ be the transformed multiset in birth-persistence coordinates~\footnote{In our experiments, we exclude points that correspond to features with infinite persistence since they are less informative than features with a defined birth and death time.}. Set $\phi_\mu:\R^2 \rightarrow \R$ be a differentiable probability distribution  with mean $\mu = (\mu_b,\mu_d) \in \R^2$ and bandwidth $\sigma$. 

The corresponding  \emph{persistence surface} is a function 
$\Phi: \R^2
\rightarrow \R$ defined by
 $\Phi(T(D)) = \sum_{\mu \in T(D)} \mathbf{w}(\mu) \phi_\mu(z)\;$ for any $z \in \R^2$, where
$\phi_\mu(\cdot)$ is the Gaussian kernel function as described above. $\mathbf{w(\cdot)}$
is a weight function, which is typically a piecewise linear function.  The \emph{persistence image} \cite{adams2017persistence} is obtained by discretizing ~$\Phi(T(D))$ and taking samples over a fixed regular grid. To be precise, we choose a rectangular region in the plane with a collection of $n\times n$ pixels, and compute the value of each pixel over the region within the bounding box of the interval by $I(D) := \iint \Phi(T(D)) dydx$, where~$x$ and~$y$ are the direction of the grid. The resulting image is denoted as $I(D)$. For simplicity, we drop the function notation and refer to a
persistence image as just $I$.

In the original paper, the weight function $\mathbf{w(\cdot)}$ is defined as the persistence of a feature. Persistence images with such a weight, we refer to as {\em persistence-weighted
persistence images}. This weight function is also commonly utilized in other proposed methods for vectorizing topological features. As previously mentioned, persistence may not always be the appropriate weight.
To enable a more flexible
weight function, Divol et al.~\cite{divol2019density} first proposed a cross-validation method to
select a better weight function of persistence images for different datasets,
their result showed customized weight function for each dataset leads to better
accuracy when using topological representation in classification. 

In this work, we also do not assume persistence is the measure of importance but
build models to learn the correct weight. Similar work has also been pursued by
Zhao et al.~\cite{zhao2019learning} where they proposed a kernel method to
learn a similarity metric for persistence images based on class labels. The
learned metric on persistence images is then applied to graph classification.
However, this work only investigated a non-deep distance metric of topological features
without consideration of interpreting the importance of topological features. 
In contrast, we propose a deep metric learning model, which combines a deep neural
network and metric learning. As we show, our deep network approach outperforms
this previous work concerning classification accuracy.  More importantly, 
using a deep metric allows explainable deep learning approaches
to extract the importance of topological features used in the classification. We use this importance to provide, for the first time, a visualization of what topological features define a class. 


\subsection{Visualizing Topological Features}
\label{sec:vis_back}

Persistence diagrams are specialized scatter plots, therefore their
visualization is straightforward and generally has not changed from their
inception.  The majority of work on visualizing topological features has focused
on features that have a direct geometric interpretation. For instance, it is
common to visualize manifolds and cells of uniform gradient flow in a
Morse-Smale complex or the branching structures of contour
trees~\cite{carr2003computing} and Reeb
graphs~\cite{biasotti2008reeb,pascucci2007robust}. There has also been working to
visualize the generators from homology
groups~\cite{obayashi2018volume,iuricich2021persistence}
(or cohomology groups~\cite{wang2011branching}) as a way of aiding the analysis of data.  Finally,
systems~\cite{obayashi2021persistent,tierny2017topology} for topological
analysis allow the visualization of the topological features (critical point
pairs) embedded directly in the scalar fields that produce them. As mentioned
previously, persistence is often the default measure of importance.
Therefore, visualizations produced by users of these systems commonly color or resize
these pairs based on
persistence~\cite{kontak2019statistical,favelier2018persistence}. In this work,
we provide the first approach to visualize a proxy for the actual {\em
importance} of topological features in classification.  In addition, we show how
our work can drive in-image visualizations with an approach to illustrating
the importance of 0D features of sublevel-set filtrations.

\section{Learning and Visualizing Topological Importance}

As we discussed \secref{bg}, persistence as a weight for importance is not
the best choice for some applications.  Rather than assume that importance can
be guessed in advance, it is better to build an approach that learns the best
weight for topological features. Since we need a basis to learn these weights,
we restrict our approach to the classification of known and unknown class
labels. A learned weight function will also provide insight into which
topological features are important in determining class label. To accomplish our
goal, we propose a deep metric model using a convolutional neural network (CNN)
with an attention module. After this model is trained, we utilize explainable
machine learning techniques to visualize the importance of topological features.
At a high level, our approach has two parts: learning a weight on topological
features in \secref{ml} and visualizing the learned weight in \secref{vispd}.


\subsection{Metric Learning for Topological Classification}
\label{sec:ml}

We use persistence images~\cite{adams2017persistence} as initial vectorized 
density estimators of diagram points. Rather than use the typical persistence weights, we use a
uniform weight, $\mathbf{w}(\cdot) = 1$. This allows our CNN to learn how to
re-weight the pixels of these {\em unweighted} persistence images such that
classes are well-separated.

\begin{figure}[t]
  \centering
    \includegraphics[width=1.0\columnwidth]{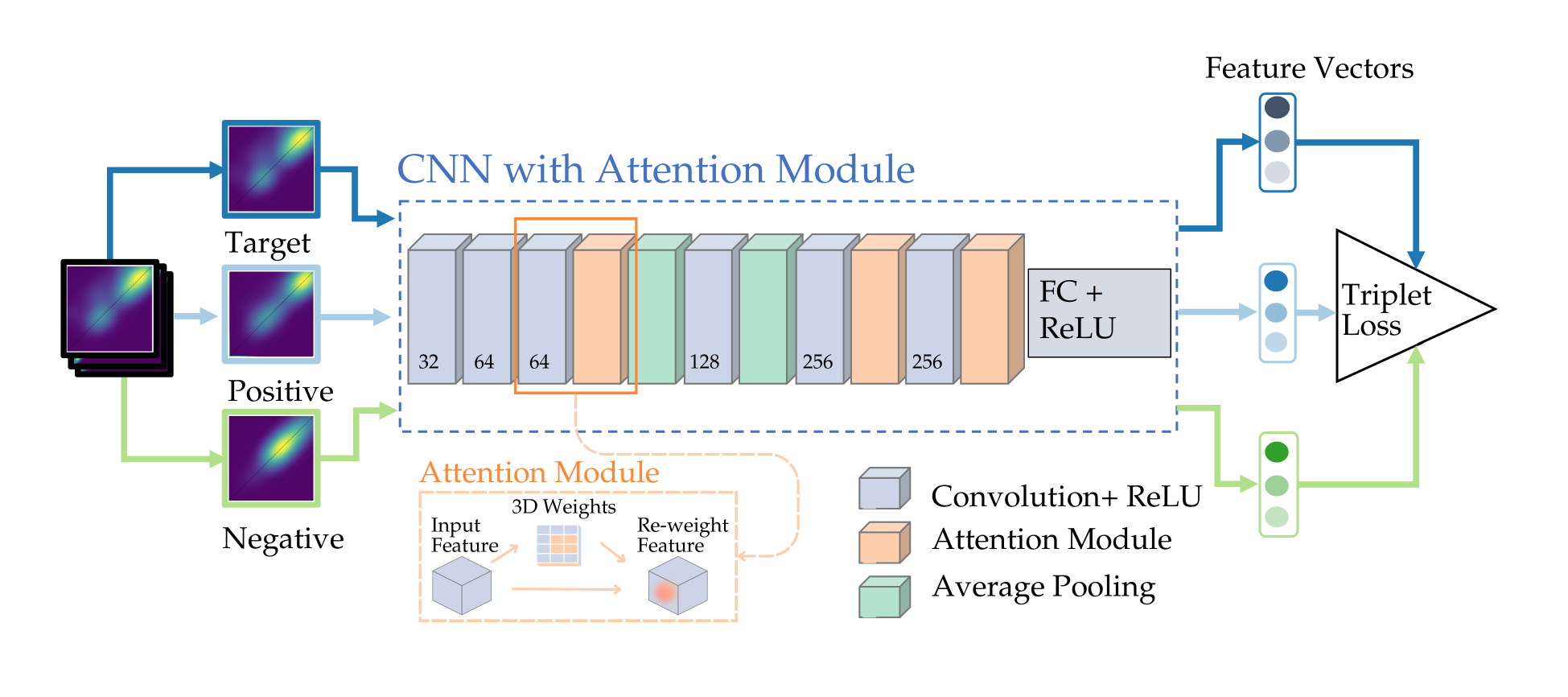}
    \vspace{-24pt}
  \caption{The architecture of our deep metric  model includes a CNN with attention modules and a metric loss function (triplet loss), where input is unweighted persistence images. The number at the bottom of a layer denotes the number of channels. FC means fully-connected layer and rectified linear activation function (ReLU) is a piecewise linear function. In training, a target is chosen with a randomly sampled positive and negative example.}
  \label{fig:model}
      \vspace{-18pt}
\end{figure}

To achieve this goal, we introduce our deep metric learning framework as shown
in \figref{model} that contains the following modules: a CNN with a metric
learning loss function as described in \secref{dml} and an attention module as
outlined in \secref{am}.

\subsubsection{Deep Metric Learning}
\label{sec:dml}
Here we give a more concrete overview of the deep metric model used in this work. Given a set of labeled unweighted persistence images, the goal is to learn a weight that can distinguish between similar and dissimilar samples. This learned weight is used as the basis for our visualization of topological feature importance. Our model uses a deep neural network to learn a feature vector and then uses a metric loss function to learn a similarity metric based on these features.

We tested two potential CNNs for our deep metric model: one standard CNN and VGG16 \cite{simonyan2014very} containing 13 convolutional layers. For both of our CNN architectures, we applied an attention module (see \secref{am} ) for refinement. In our testing, we found that the feature vectors produced for unseen data by the standard CNN were slightly more accurate (+1\%) than VGG16. Therefore, our deep metric model uses the smaller, 6 convolutional layered CNN with one fully-connected layer as shown in \figref{model}.


We use triplet loss as the metric loss function in our model due to its aptitude for learning meaningful feature representations. Triplet loss excels in comparing instances, making it ideal for capturing topological structures. By utilizing anchor, positive, and negative examples, it guides the model to create embeddings that respect data topology. This aligns with our goal of visualizing and classifying topological features. Additionally, triplet loss enables us to integrate domain-specific knowledge by selecting instances strategically, enhancing interpretability and performance.

\textbf{Triplet Loss.}  This loss is computed using three input examples, chosen at random: 1) a target image $I_T$; 2) a positive example $I_P$ that has the same class label as the target; and 3) a negative example $I_N$ that has different class label as the target. Following the previous work \cite{hoffer2015deep}, the triplet loss function $L(\cdot)$ can be formulated as:
\vspace{-12pt}
\begin{multline*}
L(I_T,I_P,I_N):=
max(||f(I_T)-f(I_P)||^2-\\||f(I_T)-f(I_N)||^2+\beta,0),
\end{multline*}
where $f(\cdot)$ is the learned weight function of the deep learning model and $\beta$ is the margin for the loss, which sets the minimum distance between positive and negative examples. In the training, positive and negative examples are randomly sampled, given a target image.

\subsubsection{Attention Module}
\label{sec:am}

In order to improve the learned weight of our model, an attention module is applied to re-weight the activation map of the CNN, which gives {\em attentional} importance to each neuron. An activation map in a CNN is a 2D representation of the output of a specific layer in the network. It shows the level of activation of each neuron in the layer. The attention module integrated into CNN enables the network to assign different weights to various regions of the activation map, allowing it to concentrate on the most informative areas that were crucial in determining the final classification decision. Attentional importance is inspired by visual neuroscience where the most informative neurons suppress the activities of the surrounding neurons. This concept is applied to our CNN through an energy function $e$ that calculates the linear separability between a target neuron and others to estimate the importance of individual neurons. See \cite{yang2021simam} for a more detailed description of the energy function and approximate solution. The energy function enhances our learned weights and visualization by determining the importance of each neuron and re-weighting them accordingly. Specifically, in our testing, we observed that using this function led to a higher classification result (+3\%) compared to not using it.


Given an activation map $A \in \R^{C\times H \times W}$, where $C$ is the number of channels and $W$, $H$ are the width and height of the convolutional layer, respectively. An attention module is applied to a CNN to re-weight the activation map. The new $\hat{A}\in \R^{C\times H \times W}$ can be calculated as: \vspace{-6pt}
$$\hat{A}= sigmoid (\frac{1}{E})\odot A,$$
where $\odot$ is a scaling operator (multiplication) and $E$ aggregates all energy function $e$ values across the channel and spatial dimensions. We add this attention module to the third and last two CNN layers similar to the original paper.

\textbf{Parameter Details.}
Our deep metric learning model is trained from scratch without fine-tuning. We randomly initialize the model's weights to fully explore its parameter space without the biases or constraints imposed by a pre-existing model. The model inputs are unweighted persistence images with the size of $40\times40$ and $\sigma=0.1$, which are the same parameters used in \cite{zhao2019learning}. Both the ordinary and extended persistence diagrams can be used to generate persistence images for our input. To train our deep metric model, we set the learning rate as 0.001 and batch size as 64. Adam optimizer is used to speed up the gradient calculation and the dropout regularization method is also used to avoid over-fitting. The Rectified Linear Unit (ReLU) function is used as our activation function, $max(0,x)$, where $x$ is the value of the activation map, because we are only interested in features that have a positive impact on the class label. We use a standard setting for the triplet loss hyperparameters: a margin of 0.1 and cosine similarity distance to measure the distance between examples in the embedding space. An $L_p$ regularizer term is applied in the triplet loss calculation. For the attention module, we use the same parameter setting as \cite{yang2021simam}. Our implementation is based on PyTorch.

\subsection{Visualization of Topological Importance}
\label{sec:vispd}

As a direct benefit of using a CNN in our deep metric model, we provide the first approach to visualize the importance of topological features in classification. Note that this approach can be applied to not only our seen, training data, but also any unseen, new data. In particular, we leverage an explainable CNN method to highlight regions in our input persistence images that  contribute the most to the model's decision-making. In this section, we introduce the explainable CNN technique used by our approach, Grad-CAM~\cite{selvaraju2017grad} and how it can be used to create a field describing the importance of topological features.  This importance field can be visualized directly, mapped back to the original points in the persistence diagram, or even mapped to features in the original data as shown below.

\subsubsection{Field of Topological Importance}
\label{sec:cam}
To visualize the learned weight function in our model such that the most significant regions of topological features in the persistence image are highlighted, we apply the Grad-CAM method in the last convolutional layer. The last is chosen as its activation maps are the most meaningful as it combines information from all other layers.

\textbf{Grad-CAM} \cite{selvaraju2017grad} Given our {\em attentionally} weighted activation maps $\hat{A} \in \R^{C\times H \times W}$, where $C$ is the number of channels and $W$, $H$ are the width and height of the convolutional layer, respectively. $A^c \subseteq \hat{A}$ refers to the activation map in the $c$-th channel produced by the last convolutional layer. We first calculate the gradient of class label score for $y^k$ as $\frac{dy^k}{dA^c}$, where $k$ is the class label, $y^k$ is the predicted probability of $k$ given by the network.

Then the average gradient of the class label score, $\alpha$, can be computed under global-average-pooling as:
$$\alpha_c^k = \frac{1}{M} \sum_{i=1}^{W} \sum_{j=1}^{H}\frac{dy^k}{dA^c_{ij}},$$
where $i$ and $j$ are the index of width $W$ and height $H$, $A^c_{ij}$ is the activation weight at location ($i,j$) of the activation map, $A^c$, and $M=H \times W$. Finally, we can weigh the activation maps across all channels in the CNN through a linear combination with ReLU:
$$\text{ReLU}(\sum_{c=1}^C \alpha_c^k \cdot A^c),$$
where the ReLU function is added to filter out negative influences on the pixel of interest.

{\em The resulting weighted activation map provides a  field over the space of a persistence image that defines regions of importance.} This field can then be used as a proxy to define the importance of topological features. The following paragraphs will describe how to design visualizations of this field, including how it can be used to drive an in-image visualization of pixel importance for sublevel set filtrations.


\subsubsection{Visualization of Importance Field.} 
We can visualize the importance field of topological features by directly displaying the weighted activation map as a colored  (\textit{magma}) heatmap, similar to how it is done in other works on explainable CNNs. To help orient the visualization in regards to the original persistence diagram, we add the standard diagonal line on our map visualization. To give better intuition on the shape and amount of importance of each region, we overlay \textit{green} isocontours. We draw contours for three isovalues set to be $50\%,70\%$, and $90\%$ of the maximum importance weight. For example, the region of the inner contour line means the feature value in this region inside is greater than $90\%$ of the maximum importance weight. We can visualize the map directly (see \figref{3d_shape_example}) or as an overlay on the plot of the persistence diagram  (see \figref{protein_result-ind}).

\textbf{In-image Visualization of the Topological Importance.} 
To provide better insight into what topological features drive a classification, we show how our importance field can be used to design an in-image visualization of topological feature importance. We target the most interpretable dimension and filtration: 0D features via a sublevel set filtration of an image.

In this case, there is a natural correspondence between 0D topological features and critical pairs (minima and saddles) that define them. This correspondence, combined with our importance field over the space of a diagram, can provide an intuitive visualization of what structure defines a class in original data.



Given an image and its persistence diagram, we can use the diagram to obtain critical pair information for each point $(b,d)$ in the diagram, which corresponds to a interlevel set in the image. Specifically, let $p_b$ be the sublevel set in the image corresponding to the grayscale value $b$ (the minimum point), and let $p_d$ be the sublevel set corresponding to the grayscale value $d$ (the saddle point). Each point, $(b,d)$, can be used to look up the importance directly in our importance field (discrete, but linearly interpolated). Based on these pixel values, the corresponding pixel locations can be plotted and visualized directly, say by picking each minimum for each point like previous work~\cite{soler2019ranking,favelier2018persistence,kontak2019statistical,carriere2015stable}.  Although this only provides an intuition of the extremes in a feature, not the 0D topological feature each pair represents. Therefore, for our approach, we visualize features by drawing the interlevel set between $p_b$ and $p_d$ for each pair.


We now discuss how to visualize the interlevel set of 0D topological features. To begin, we consider the sublevel set filtration for the image, which involves constructing a nested sequence of topological spaces based on the image's grayscale values. Given an image with grayscale values ranging from $0$ to $255$. A continuous function can be defined that assigns each pixel its grayscale value. The sublevel set filtration of the image is then defined as the nested sequence of sublevel sets: $p_0 \subseteq p_1 \subseteq \cdots \subseteq p_{255}$, where the sublevel sets $p_k$ correspond to the set of pixels in the image with grayscale values less than or equal to $k$. 


Given a point $(b,d)$  in the persistence diagram, the corresponding interlevel set can be determined as $p_b \subseteq \cdots \subseteq p_d$, where $p_b$ and $p_d$ are critical pairs in the image. This interlevel set captures the birth and death of a 0D topological feature in the image, providing insight into its lifetime. By visualizing the interlevel set, we aim to gain a deeper understanding of the topological features present in the image.

We color each interlevel set based on the importance value of the diagram point in our field, again, using the \textit{magma} color map. In cases where an older feature subsumes a younger feature in the filtration, we assign the same color to both features. This is because the older feature includes the 0D feature of the interlevel set of the younger feature in our extraction.


To highlight high-importance regions we process the set of persistence points, rendering their 0D features of interlevel sets, in inverse order of importance. Thus, the most important regions are in front. Given that our image data has discrete function values, there will be the potential of several sets getting the same importance value.  Since they share the same color, their relative ordering does not matter.

\section{Results}
\label{result}
In this section, we demonstrate the effectiveness of our approach in: (1) learning a metric for topological features, such that features are weighted for accurate classification and; (2) visualizing topological importance such that key structures for the classification are highlighted. The real-world datasets evaluated in our approach are detailed in \secref{data}. We begin with a study in \secref{toy} with a scenario where we assume to have prior knowledge of the meaningful importance weight function (persistence), and show that our method can learn that weight. In practice, however, this prior knowledge cannot be assumed therefore, importance must be \textit{learned}. We evaluate our learned weight on a variety datasets and provide topological classification results in \secref{accuracy}.  We compare and show these results are more accurate than other state-of-the-art approaches. Finally, we provide examples of using the importance field extracted from the learned weight to visualize topological importance in \secref{vis_results}. {\em All examples in the following figures use unseen data to our model.} Our code is available in an \href{https://osf.io/wbtrp/?view_only=e617f1e2398e4a7a8273350183a89eca}{\texttt{OSF}} repository. Our importance field is presented with a \textit{magma} colormap, but to keep our results distinct, we present all other persistence images (weighted, unweighted) using \textit{viridis}.


\subsection{Evaluation Datasets}
\label{sec:data}
We evaluate our approach on five datasets from graph, shape, and medical imaging, which includes a range of filtration functions and dimensions of topological features.

\textbf{3D Shape}\cite{sumner2004deformation} This dataset contains 6 different 3D shape classes including faces, human heads, camels, horses, cats, and elephants. There are 1,200 persistence diagrams in total with 200 persistence diagrams for each class. Diagrams of 0D features are produced using the implementation of \cite{carriere2015stable} that uses a Vietoris–Rips filtration.

\textbf{PROTEINS}\cite{borgwardt2005protein}: This graph dataset of protein molecules contains 1,113 graphs with 2 classes: enzymes and non-enzymes. Nodes of each graph are amino acids and edges connect pairs that are less than 6 Angstroms apart. 
Following \cite{zhao2019learning}, the Jaccard-index function on graph edges allows extended persistence diagrams to be computed using sublevel-set and superlevel-set filtration to extract 0D and 1D features.

\textbf{COLLAB}\cite{yanardag2015deep}: This is a graph dataset denoting scientific collaborations in High Energy Physics, Condensed Matter Physics, and Astro Physics. This set has 5,000 graphs with 3 labels that indicate the research area.  Similar to the PROTEINS dataset, extended persistence diagrams with 0D and 1D features were produced for each graph.

\textbf{Prostate Cancer}\cite{lawson2019persistent}: This set includes 5,182 region-of-interest images from hematoxylin \& eosin (H\&E) stained histological images with 3 classes, that denote the progression of cancer (Gleason score 3, 4, and 5). Persistence diagrams of 0D features were produced for each image via sublevel set filtration using the Giotto-tda library~\cite{tauzin2020giottotda}.

\textbf{Colorectal Cancer}\cite{kather_jakob_nikolas_2018_1214456}: This is a set of 1,800 region-of-interest images from H\&E stained histological images with 9 classes. Similar to the prostate images, diagrams are obtained for 0D features via sublevel set filtration using Giotto-tda library\cite{tauzin2020giottotda}.

\subsection{Learning Persistence Weights}
\label{sec:toy}
Our deep metric model is designed to learn the best weight for diagram point density for classification. We present a scenario in which persistence is the appropriate weighting for topological features, and demonstrate how our learned weight can effectively capture "persistence". To evaluate this ability, we generated two synthetic datasets, each containing diagrams of a distinct class, with one high persistence feature present in all members of that class. Additionally, each diagram contains 100 randomly generated low persistence points, representing random noise. This scenario tests the efficacy of using persistence as a measure of importance when one high-persistence feature defines a class amidst low-persistence noise.

\begin{figure}[tb]
    \centering
    \includegraphics[width=\columnwidth]{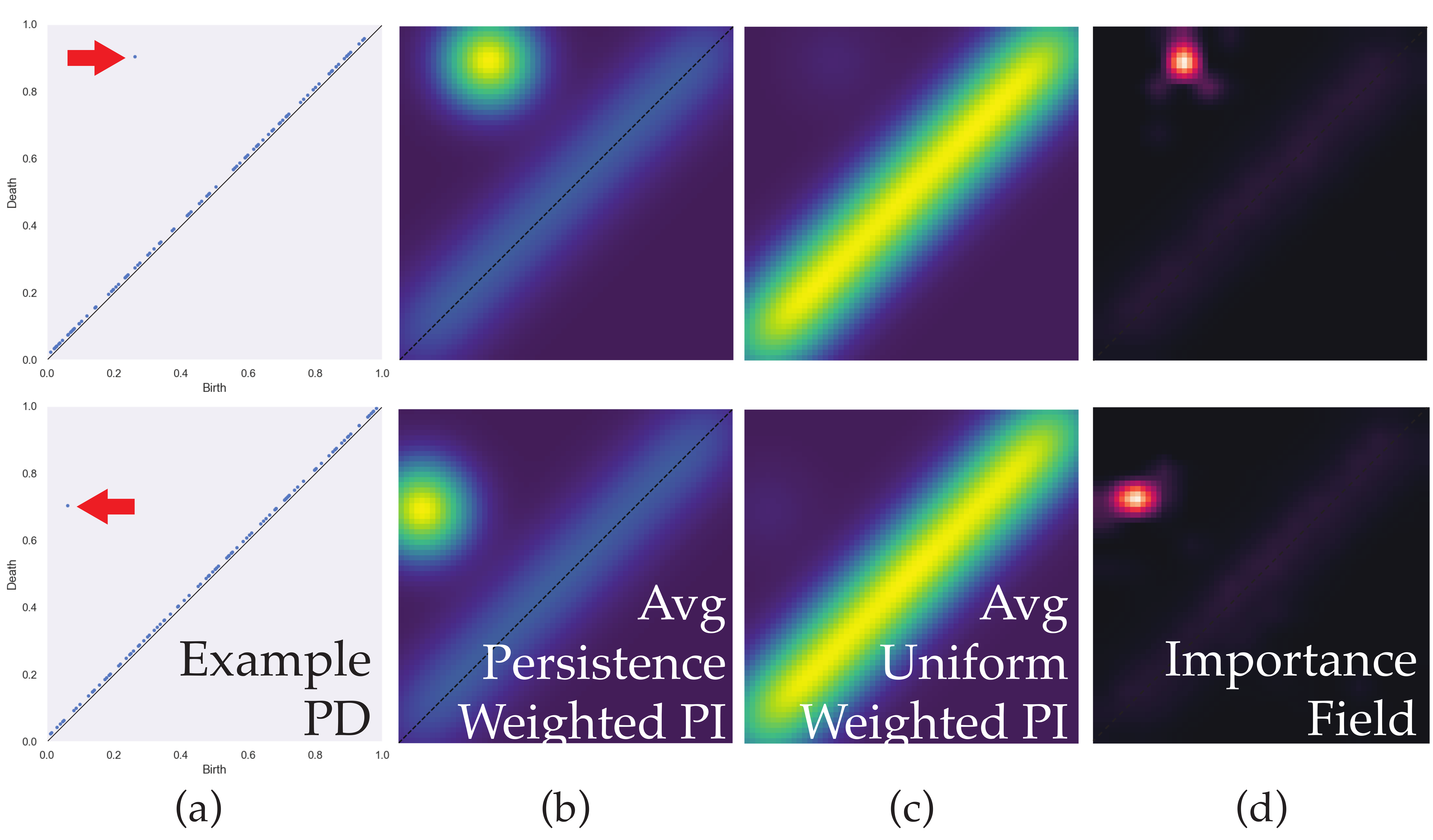}
    \caption{(a) Example persistence diagrams (PD) for 2 classes with 0D features.  Each class has one high persistence point and a random distribution of many low persistence points. In this case, a persistence weight would be ideal for classification. (b) A persistence weighted persistence image (PI). (c) Given a uniform density distribution, (d) our approach can learn to weight by persistence.
    }
    \label{fig:2_class_example}
    \vspace{-10pt}
\end{figure}

\figref{2_class_example} illustrates our results. \figref{2_class_example} (a) gives an example diagram from each class where the important high persistence feature is denoted with a red arrow. \figref{2_class_example} (b) shows the average for all class members of the standard persistence-weighted persistence image where the high persistence features receive a larger weight. \figref{2_class_example} (c) is the average unweighted persistence image that gives the density of points, where all points are considered equal. This is the average image of the inputs to our approach. Note that the diagonal noise dominates as it contains the highest density of points. As persistence is the ideal weight for this scenario, if our approach works as it should, the average importance field produced should be similar to the average persistence-weighted persistence image. \figref{2_class_example} (d) shows that this is the case since high persistence features are deemed important and low persistence features are discounted. Therefore, our approach can learn persistence weighting if that weight is the right one for a dataset, but it is not limited to only considering that measure of importance.

\begin{figure}[tb]
    \centering
    \includegraphics[width=\columnwidth]{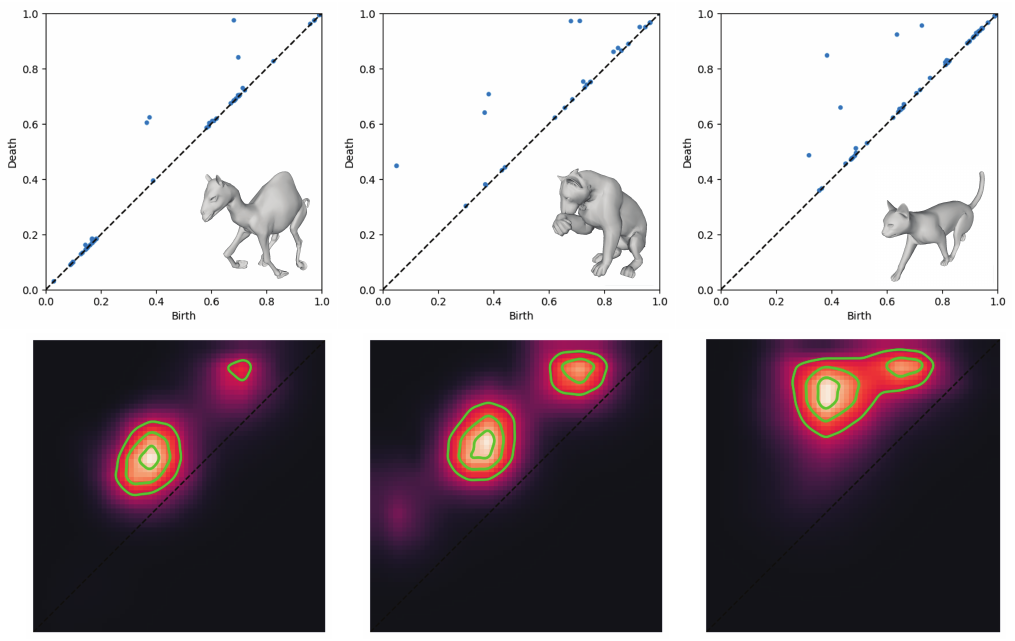}
    \caption{3D Shape examples along with a visualization of topological importance for their classification. The top portion of each figure shows the persistence diagram for the 3D shape example with 0D features, while the bottom portion shows our visualization of the importance field. In this case, our deep learning model leans a persistence-like weight of features.}
    \label{fig:3d_shape_example}
        \vspace{-12pt}
\end{figure}

We provide another, real-world example of our approach learning a persistence-based weight in \figref{3d_shape_example}. This dataset consists of 3D shapes and has been previously used for feature tracking based on high-persistence topological features \cite{carriere2015stable}. Given this prior use, we can assume that standard persistence-based weighting would yield satisfactory results. This assumption is supported by the accuracy of persistence-based weighting strategies in our classification results, which are discussed in Section \ref{sec:accuracy}. As shown in \figref{3d_shape_example}, our deep metric model indeed learns to assign importance based on persistence. High-persistence features are given more weight, while low-persistence features are discounted.

\bgroup
\addtolength{\tabcolsep}{-2pt}
\begin{table}[tb]
  \caption{Classification accuracy percentage using topological features, comparing our approach with 1-Wasserstein distance (W1), persistence-weighted persistence images (PWPI), weighted persistence image kernel (WKPI), Betti curves (BC), persistence-weighted Gaussian kernels (PWGK), and sliced Wasserstein kernel (SWK).}
  \label{tab:svm}
  \scriptsize%
	\centering%
  \tabulinesep=1.2mm
  \begin{tabu}{llllllll}
  \toprule
 Dataset &W1 &PWPI&WKPI &BC & PWGK & SWK &Ours  \\
  \hline
    	3D Shape &0.92 &0.89 &\textbf{1.0} &0.91 &0.9 &0.88 &\textbf{1.0}   \\
    	COLLAB  &0.76 &0.73 &0.77 &0.76 &0.71 &0.78 &\textbf{0.84} \\
  	PROTEINS &0.77 &0.76 &0.79 &0.76 &0.72 &0.76 &\textbf{0.87}    \\
  	Prostate &0.82 &0.85 &0.88 &0.86  &0.83 &0.84 &\textbf{0.95}   \\
  	Colorectal &0.77 &0.78 &0.81 &0.77 &0.78 &0.75 &\textbf{0.85}    \\
  \bottomrule
  \end{tabu}%
  \vspace{-12pt}
\end{table}
\egroup

\subsection{Learned Weight Accuracy}
\label{sec:accuracy}
For our visualization to be effective, the deep metric classifier on which it is based should be accurate. To this end, we evaluate the accuracy of our learned weight by comparing it to other commonly used topological representations in classification. Similar to the state-of-the-art approach in learned topological classification \cite{zhao2019learning}, we employ an SVM-kernel classifier and adopt a 90/10 training-test data split. In order to ensure a fair comparison, we use the same training and testing dataset for both our deep metric model and the classifier. Accuracy results are based on the classification of the unseen, test data.

We compare our method to other topological representations frequently used in classification, such as persistence diagrams using 1-Wasserstein distance (W1), persistence-weighted persistence images (PWPI) \cite{adams2017persistence}, Betti curves (BC) \cite{rieck2017topological}, persistence-weighted Gaussian kernels (PWGK) \cite{kusano2016persistence}, and sliced Wasserstein kernels (SWK) \cite{carriere2017sliced}. Additionally, we compare our method to the previous state-of-the-art in learned weights for topological classification: the weighted persistence image kernel (WKPI) \cite{zhao2019learning}. We use the same parameter settings for persistence images (size $40\times40$, $\sigma=0.1$) and Betti curves (BC) size ($40\times40$) as described in \cite{zhao2019learning}. The training time for our method is at least 5 times faster than the WKPI method for learning to weight, as we employ a more streamlined model. To illustrate, when considering the Colorectal dataset with 1800 datapoints, our training process takes approximately 11.25 minutes, while WKPI requires over an hour.

A sensitivity analysis was conducted to evaluate how changes in parameter settings affect the persistence images and Betti curves in Prostate cancer image classification. The size of persistence images and Betti curves were varied of the range $[10 \times 10$, $100 \times 100]$ in increments of 1, and $\sigma$ of persistence image was varied of the range $[0.001,1]$ in steps of 0.001. The results showed that the classification accuracy was not significantly affected, with a difference of less than 1\%.

The classification results in \tabref{svm} demonstrate that our approach outperforms traditional W1 distance of persistence diagrams in terms of accuracy. In fact, our approach achieves a significant improvement in accuracy over the next best method in the graph classification task for the COLLAB dataset, with an increase of +6\%. Although W1 already provides accurate results for 3D shape classification, our method further improves accuracy to achieve perfect classification results. For the prostate imagery dataset, our method achieves an increase in accuracy of +7\% over the next best method, resulting in an overall classification accuracy of 95\%. Our approach also yields a +4\% improvement in classification accuracy for the colorectal cancer dataset.

Of particular note is our accuracy for the PROTEINS graph classification (87\%). This not only outperforms the other representations (an increase of +8\% compared to the next best), it outperforms the best-known machine learning approach~\cite{zhang2019hierarchical} (85\%) according to the~\href{https://paperswithcode.com/sota/graph-classification-on-proteins}{\texttt{Papers with Code}} website at the time of submission.

Our results illustrate how our learned weight outperforms approaches that assume persistence as the measure of importance (PWPI, BC, PWGK, SWK). This implies that persistence is not the ideal weight in these datasets. In addition, our approach is comparable or better than the state-of-the-art in learned weights for topological classification, WKPI. This indicates that our approach's use of a deep learning network is more effective than the previous work.

The accuracy of our classification results motivates our next step: visualizing the importance of topological features in order to understand what topological features are key to defining classes.

\subsection{Visualization Results}
\label{sec:vis_results}

In this section, we present both the visualizations of our proposed topological importance field and an in-image visualization of the topological significance. The former allows for a clear representation of the learned topological importance, while the latter highlights the significance of topological features within the data itself.

We demonstrate the effectiveness of our visualization by answering the following questions: (1) Can our visualization show different importance regions in different classes while persistence cannot?; (2) Can our visualization indicate similar topological importance regions in the same class while persistence cannot?; and (3) Can our visualization show examples with varied importance regions but with similar and meaningful structures in the data itself?

\begin{figure}[tb]
    \centering
    \includegraphics[width=0.9\columnwidth]{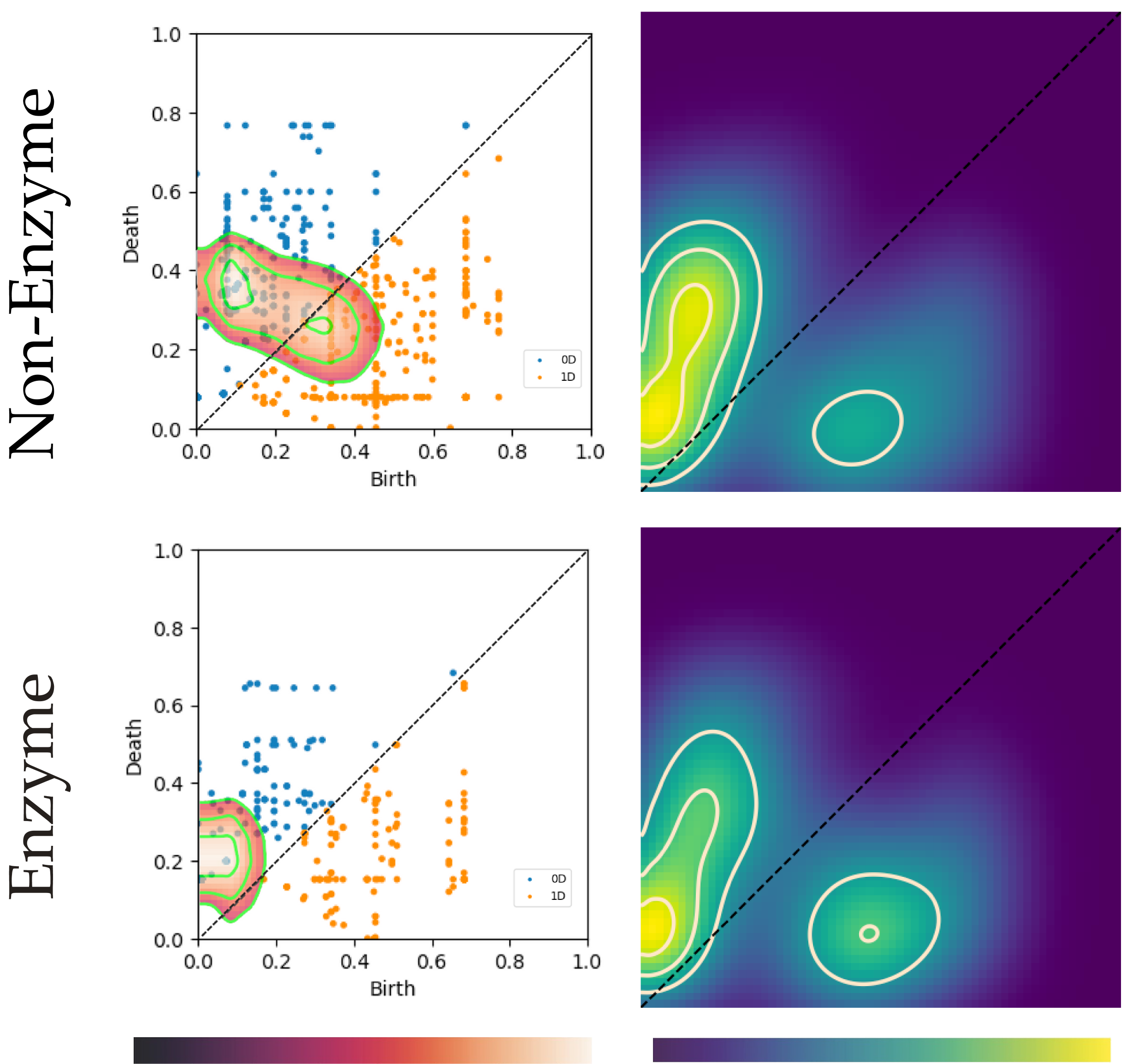}
        \vspace{-8pt}
    \caption{
    PROTEINS graph classification with two classes: enzymes and non-enzymes. (left) Example persistence diagrams with 0D and 1D features overlaid with our visualization of the importance field for each class, which has the smallest W1 distance. (right) Visualization of the corresponding persistence-weighted persistence images.
    }
        \vspace{-18pt}
    \label{fig:protein_result-ind}
\end{figure}

The importance field visualization for PROTEIN answers the first question in \figref{protein_result-ind}, providing the importance field obtained from the learned weight and persistence weight of example classes: non-enzyme and enzyme. We randomly selected an enzyme example from the testing data, and the non-enzyme example was obtained by calculating the 1-Wasserstein (W1) distance across the testing data, which had the smallest W1 distance compared with the enzyme example. In other words, a traditional uniform, persistence-based approach would not consider these two datasets to be in separate classes. Our results show that their persistence-weighted persistence images look virtually identical. Hence, the pre-assumed and fixed weight function (persistence) is not an appropriate metric to distinguish these two examples. However, our visualization of the importance field indicates a significant difference between the two classes. Specifically, the non-enzyme example highlights the importance of both 0D and 1D features in classification, with a mix of low and medium persistence 0D features and low persistence 1D features. In contrast, in the enzyme example, the 0D features are born at low function values and are the most important in classification, with a negligible contribution from 1D features.

\begin{figure}[tb]
    \centering
    \includegraphics[width=0.9\columnwidth]{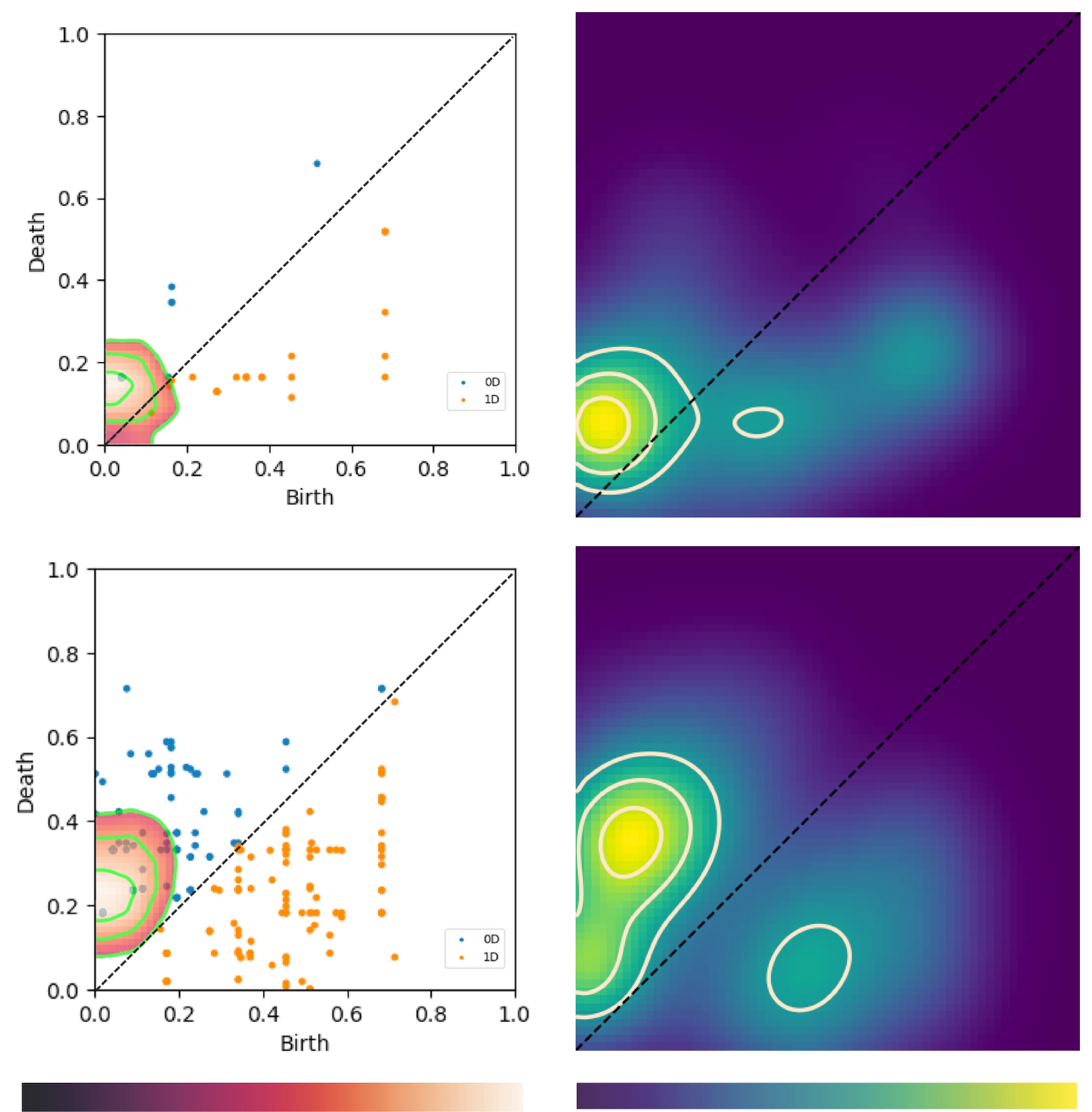}
        \vspace{-8pt}
    \caption{
    PROTEINS graph classification with class enzymes. (left) Example persistence diagrams with 0D and 1D features overlaid with our visualization of the importance field for same class, which has the largest W1 distance. (right) Visualization of the corresponding persistence-weighted persistence images.
    }
        \vspace{-18pt}
    \label{fig:protein_result_inclass}
\end{figure}

To attempt to answer the second question, we used the PROTEIN dataset as evidence, as shown in \figref{protein_result_inclass}. This result presents the importance field obtained from our learned weight and persistence weight of the same class, enzyme, where these two examples have the largest W1 distance. In other words, we found a dataset where a traditional persistence-based weight would likely treat these two datasets as being from separate classes. This is further illustrated by the two examples showing notable differences in their persistence-weighted persistence images.
These differences indicate that the persistence weight function alone is an insufficient metric for classification. 

However, our importance field visualization highlights that the most important features for classification are the 0D features that are born at the low function value, which are similar in both examples. Therefore, our \textit{learned} metric is superior for classification, and the evaluation results in \tabref{svm} support this conclusion.

\begin{figure}[tb]
    \centering
    \includegraphics[width=0.9\columnwidth]{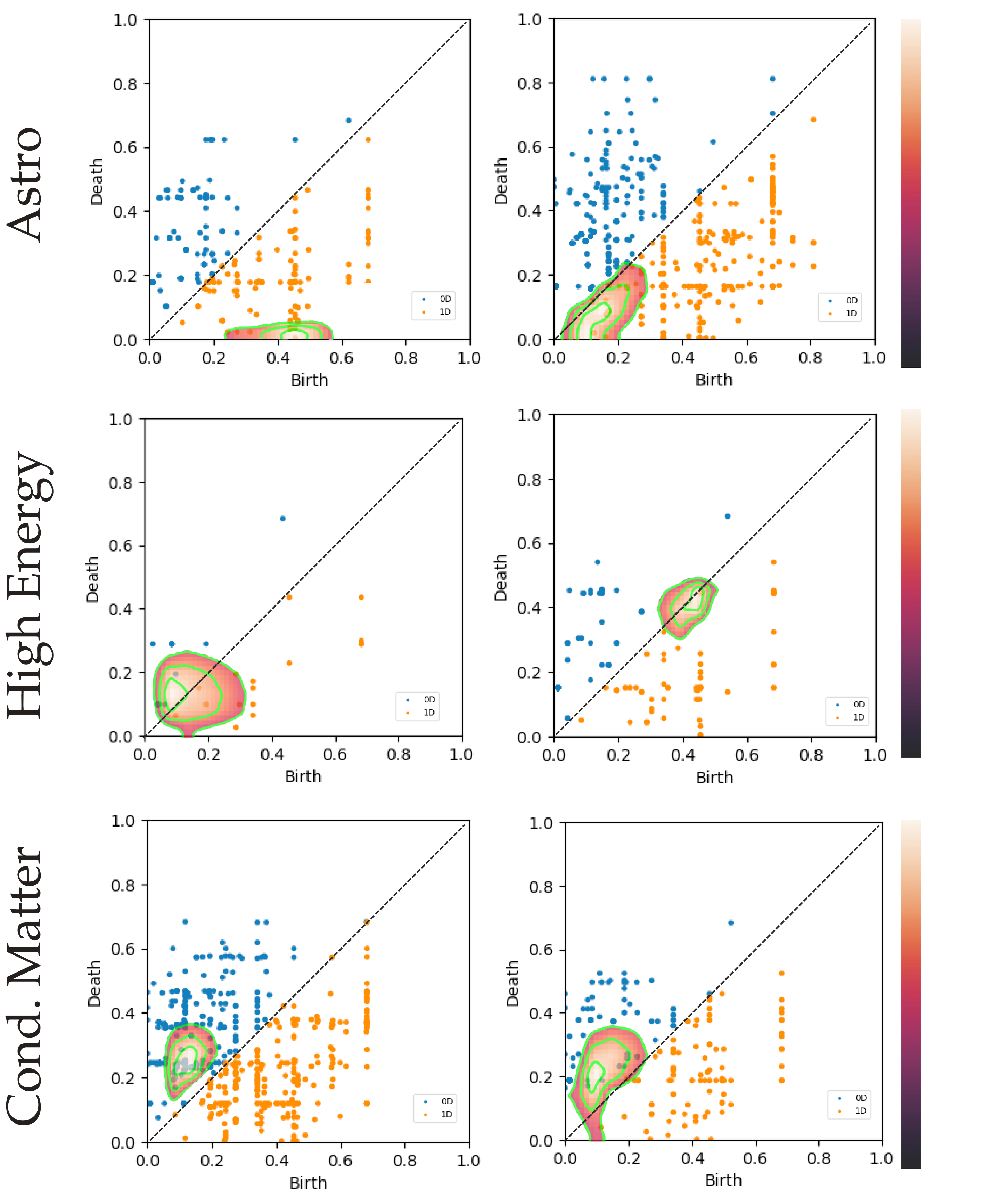}
    \vspace{-10pt}
    \caption{COLLAB graph classification with three classes. (left) Example persistence diagrams with 0D and 1D features overlaid with our visualization of the importance field from 3 collaboration networks in physics. (right) Example persistence diagrams overlaid with our visualization of the importance field for the same class. 
    }
    \label{fig:collab_result_ind}
    \vspace{-18pt}
\end{figure}
\figref{collab_result_ind} shows the importance field visualization for graph examples from scientific collaboration networks in High Energy Physics, Condensed Matter Physics, and Astro Physics (COLLAB). For each class, the plots show two example persistence diagrams from each class overlaid with the importance field visualization.

For High Energy Physics in \figref{collab_result_ind}, the first example demonstrates that the classification is determined by both low persistence 0D and 1D features that are born and die at low function values. The second example highlights the significance of both low persistence 0D and 1D features as well, while they are born and die at medium function values. In Condensed Matter Physics, both examples have a similar importance field, with low-medium persistence 0D features born at low-medium function values being the most significant. In Astro Physics, both examples show that the 1D features (lower triangle in the diagram) are important in classification, despite being in different ranges of persistence and birth-death locations. 
This implies that collaboration loops between authors are likely more indicative in Astro Physics than other classes (i.e., Condensed Matter Physics). These examples illustrate the use of a mix of high, medium, and low persistence features in classification, suggesting that a single weighting scheme (persistence, inverse-persistence, or other) would not yield high-quality classifications. This is supported by our quality measures in \tabref{svm}.

\begin{figure*}[tb]
    \centering
    \includegraphics[width=\linewidth]{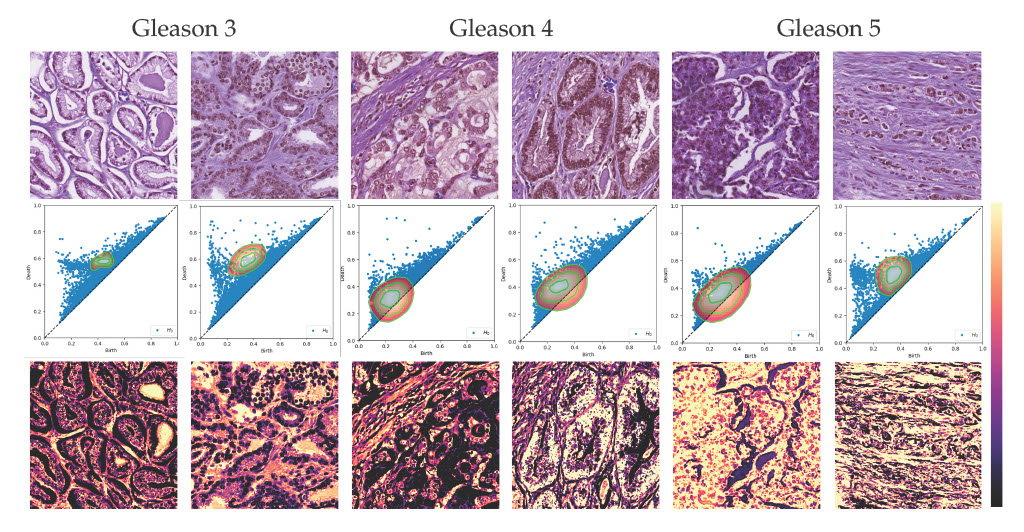}
    \vspace{-24pt}
    \caption{Prostate cancer medical image classification. Each class has two examples in Gleason 3,4 and 5. From top to bottom are the original image, its persistence diagram with 0D features overlaid with our learned importance, and in-image visualization of topological importance.
    }
    \label{fig:new_prostate_result_ind}
    \vspace{-14pt}
 \end{figure*}

\begin{figure*}[t]
    \centering\includegraphics[width=\linewidth]{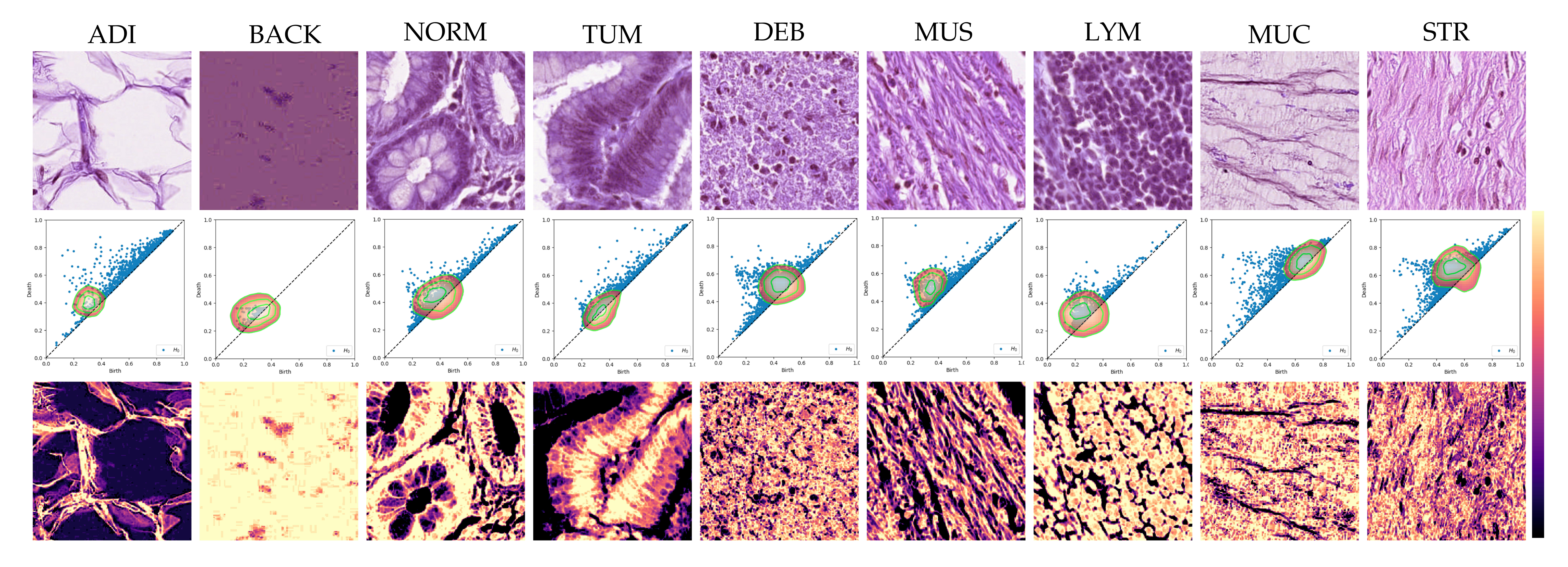}
    \vspace{-24pt}
    \caption{Colorectal cancer image examples from 9 classes. Each column is an example image per class, its persistence diagram with 0D features overlaid with our learned importance, and in-image visualization of topological importance.}
       \vspace{-18pt}
    \label{fig:Colorectal_column}
\end{figure*}

Our visualization results for medical imaging datasets provide an answer to the third question by demonstrating the in-image visualization of topological importance, which effectively highlights the medically significant structures within the data. This visualization is interpreted by our biomedical collaborators. 

In \figref{teaser} and \figref{new_prostate_result_ind}, we present the visualization results for digital pathology images of prostate cancer. The dataset includes examples of Gleason 3 to 5, where higher grades correspond to more advanced stages of the disease.  A hallmark of this disease is that well-formed prostate glands deteriorate and lose structure, such that at more advanced stages, no glandular structure is present. As our visualization shows, this is indeed the case as the stroma defining the glandular structure are the most important features in the classification of Gleason 3. As cancer progresses to Gleason 4 and as the glands break down, the important features become more cellular, involving both the semi-structural stroma and nuclei. At the final stage of progression in Gleason 5 where glands have entirely deteriorated, no structure is present, and the important features become local/cellular information. These important features are not captured by common measures of importance, such as persistence, and can be identified and visualized for the first time with our approach. For example, prostate calcifications only occur in well-formed glands.  In \figref{teaser}, these are important features in our field (red arrow).
Furthermore, as \figref{new_prostate_result_ind} illustrates, important regions in the diagram vary in different examples, while our in-image visualizations show that they correspond to similar structures. This provides further evidence that a single weighting strategy is not ideal for this dataset.  This is supported by the fact that our learned weight achieves 95\% accuracy, as shown in \tabref{svm}.
%



Our final example, shown in \figref{Colorectal_column}, demonstrates the application of our method to colorectal cancer image classification with 9 classes. Each class has various of structural arrangements and distributions. Our visualization provides the first step to interpret the structural difference between them by highlighting the importance of topological features in distinguishing their classes. For instance, normal colon mucosa (NORM) features a uniform and regular arrangement of epithelial cells with glandular structure, while cancer-associated stroma (STR) displays a more disorganized arrangement that disrupts the normal tissue structure. As shown in this figure, our visualization emphasizes such structural importance. Moreover, in comparison with NORM
class, colorectal adenocarcinoma epithelium (TUM) also has glandular
structure, but with abnormal epithelial cells. Our in-image visualization
highlights the glandular structure in both NORM and TUM, indicating
the normal cells in NORM and the abnormal cells in TUM.

Other examples further demonstrate how our visualization aligns with the important structural characteristics of each class. For instance, the background (BACK) example shows that the important features are mainly artifacts, indicating that this class is predominantly noise. Adipose (ADI) tissue comprises adipocytes and connective tissue, and our in-image visualization highlights the connective tissue structure. Debris (DEB) refers to damaged tissue that has broken down, which is also reflected in the lack of structure in the in-image visualization. Next, our importance visualization highlights the fibers of found in smooth muscle (MUS) and the sparse tissue of mucus (MUC). Lymphocytes (LYM) are a type of white blood cell, and our in-image visualization emphasizes that the cell structure is crucial in determining class.

All persistence diagrams overlaid with our learned importance in this example are distinct, indicating the need to learn the weight function instead of relying on a pre-fixed one (i.e. persistence). This example demonstrates how our visualization can effectively highlight the medically significant structures in a complex dataset, a conclusion supported by feedback from our biomedical collaborators.
\section{Discussion}
In this paper, we introduced the first visualization of the importance of topological features, which includes the visualization of an importance field through the learned weight function and in-image visualization of topological significance. This allows TDA researchers to gain insight into the topological features that drive dataset classification for the first time. Rather than an assumed, fixed weighting our novel deep metric model optimizes the weight function given labeled data. Furthermore, our model outperforms other topological representations, including those that use persistence-based weights or learned kernel weight functions.  

However, our novel approach also has limitations. The persistence image, acting as a density estimator, may lead to density and importance overlapping along the diagonal due to smoothing. This could pose challenges for points near the diagonal on extended diagrams. We demonstrated how our field can drive in-image visualization of 0D features through sublevel set filtrations on images. While our field is dimension and filtration agnostic, mapping diagram points back to original data is not always straightforward. For instance, visualizing generators for even 1D features remains an active research area~\cite{iuricich2021persistence,obayashi2018volume,li2021minimal}. Additionally, visualizing importance in unstructured datasets is an open question. These topics offer exciting avenues for future research. For example, our approach could extend ongoing work on visualizing 1D features or enhance visualizations of the Morse-Smale Complex.

Our results highlight the variability of topological importance across domains, classes, and datasets. Thus, a single fixed weighting strategy may not be optimal for various datasets. This emphasizes the need for further research in this domain, where our approach serves as the pioneering visualization tool.

\acknowledgments{%
This work has received support from the Department of Energy, the National Science Foundation, and the National Institutes of Health (DOE ASCR DE-SC0022873, NSF-IIS 2136744, NIH R01GM143789, NSF CCF 2046730, and NSF DMS 1664858).
}

\bibliographystyle{misc/abbrv-doi-hyperref}

\bibliography{paper}

\end{document}